\documentclass[conference]{IEEEtran}
\IEEEoverridecommandlockouts
\usepackage{cite}
\usepackage{subcaption}
\usepackage{amsmath,amssymb,amsfonts}
\usepackage{algorithmic}
\usepackage{graphicx}
\usepackage{textcomp}
\usepackage{xcolor}
\usepackage{epstopdf}
\def\BibTeX{{\rm B\kern-.05em{\sc i\kern-.025em b}\kern-.08em
    T\kern-.1667em\lower.7ex\hbox{E}\kern-.125emX}}
\begin{document}

\title{AgroXAI: Explainable AI-Driven Crop Recommendation System for Agriculture 4.0
}

\author{
\IEEEauthorblockN{Özlem TURGUT}
\IEEEauthorblockA{Department of Computer Engineering\\
Pamukkale University\\
Denizli, Turkey\\
Email: oturgut19@posta.pau.edu.tr}
\and
\IEEEauthorblockN{İbrahim KÖK}
\IEEEauthorblockA{Department of AI and Data Engineering\\
Ankara University\\
Ankara, Turkey\\
Email: ikok@ankara.edu.tr}
\and
\IEEEauthorblockN{Suat ÖZDEMİR}
\IEEEauthorblockA{Department of Computer Engineering\\
Hacettepe University\\
Ankara, Turkey\\
Email: ozdemir@cs.hacettepe.edu.tr}
}

\maketitle

\begin{abstract}
Today, crop diversification in agriculture is a critical issue to meet the increasing demand for food and to improve food safety and quality. This issue is considered to be the most important challenge for the next generation of agriculture due to diminishing natural resources, limited arable land and unpredictable climatic conditions caused by climate change. In this paper, we employ emerging technologies such as the Internet of Things (IoT), machine learning (ML) and explainable artificial intelligence (XAI) to improve operational efficiency and productivity in the agricultural sector. Specifically, we propose an edge computing-based explainable crop recommendation system, AgroXAI, which suggests suitable crops for a region based on weather and soil conditions. In this system, we provide local and global explanations of ML model decisions with methods such as ELI5, LIME, SHAP, which we integrate into ML models. More importantly, we provide regional alternative crop recommendations with the Counterfactual explainability method. In this way, we envision that our proposed AgroXAI system will be a platform that provides regional crop diversity in the next generation agriculture. 
\end{abstract}


\section{Introduction}
Today, global climate change, population growth, agricultural land depletion and biodiversity loss pose serious environmental, social and economic threats \cite{friha2021internet}. Under these globally changing conditions, industrial growth and production methods based on sustainable agricultural principles are needed to meet the increasing demand for food and improve food security and quality \cite{ccakmakcci2023assessment}. At this point, ensuring crop diversity in agriculture stands out as a critical issue. Crop diversification is the process of adding a new crop, a different variety, to a field \cite{rahman2024crop}. It is not only mitigates various agricultural challenges such as soil degradation, pest infestation, and climate change impacts, but also enhances farm resilience, fosters income growth on small holdings, stabilizes commodity prices, improves soil health, and offers a diverse range of nutritious food options for both humans and livestock \cite{barman2022crop}. In this context, ensuring regional crop diversity based on low-cost digital technologies, in response to changing environmental and climate conditions, emerges as a key area of focus for agriculture 4.0 \cite{bui2024agriculture}. In particular, it is necessary to periodically monitor the soil, air, and environmental factors in distributed geographic regions and provide suitable crop alternatives accordingly.

The concept of the Internet of Things (IoT), which provides technology-oriented creative solutions to the current needs in the new agricultural era, comes to the forefront \cite{alam2023analysis,alzubi2023artificial}. IoT offers innovative digital solutions in many areas based on low-cost, low-latency approaches such as edge and fog computing that respond to end-user needs on-site \cite{kok2019deep}. Many of these solutions are also being used effectively in agriculture domain. In this context, a wide variety of IoT-supported smart agriculture applications are implemented in crop management, soil management, water management, livestock management, green management, weather management, and tracking and tracing \cite{sinha2022recent,ccetin2022smart}. It is observed that these applications are becoming widespread with AI support, but there is also a growing need for interpretability and explainability for the effective use of these technologies. In this regard, Explainable Artificial Intelligence (XAI) plays a crucial role in agriculture by enhancing the transparency of AI-driven recommendations, such as those for crop selection, soil health, and pest management \cite{kok2023explainable}. By clarifying how environmental factors, soil properties, and climate conditions shape AI suggestions, XAI builds trust in technological solutions and supports data-driven decision-making. This interpretability is essential for sustainable agricultural practices, allowing stakeholders to make informed choices that enhance resilience and productivity amid changing environmental conditions.

Specifically, the existing research efforts related to crop prediction in the agriculture domain are as follows.

Sharma et al. \cite{sharma2021ai-farm} developed a crop prediction model based on Gaussian Naive Bayes, Decision Tree, Logistic Regression, Random Forest, and XGBoost models. The authors developed their models by training them on data collected from different agricultural fields in India. The forecasting system based on the developed models was enabled to be used online and offline via an Android application.  
Similarly, Doshi et al. \cite{doshi2018agroconsultant} developed an intelligent recommendation system called AgroConsultant to help farmers in India make an intelligent decision on which crop to grow depending on various environmental and geographical factors. In the developed system, the authors presented a system design that combines data sets showing soil, climate and precipitation characteristics and decides which crop farmers will grow by using Decision Tree, K-Nearest Neighbour, Random Forest, regression and Neural Network models. 

Rajak et al. \cite{rajak2017crop} aimed to design a crop recommendation system using an ensemble learning model based on Support Vector Machine, Naive Bayes, Multi-layer Perceptron and Random Forest models. The proposed system was tested with soil dataset collected from different sources. Rule-based output was produced with the ensemble model used in the system. Similarly, Pudumalar et al. \cite{pudumalar2017crop} aimed to design a highly accurate and effective crop recommendation system using ensemble techniques. They applied the majority voting technique in Random Forest, CHAID, K-Nearest Neighbour and Naive Bayes learners. In this study, soil testing lab and crop data available in online sources were used as a dataset. As a result of the applied model, they created a rule. Saranya et al. \cite{Saranya_K_2023} developed a crop recommendation module that ensembles the results of Support Vector Machine, Random Forest, Naive Bayes, and K-Nearest Neighbor models using majority voting. In addition, they designed a web-based interface called TILLAGE that includes fertilizer recommendation and pesticide recommendation modules based on the CNN model. The proposed modular solution aims to increase agricultural production by improving soil use, fertilization and crop selection. 

Unlike the above studies, edge computing architecture and Deep Reinforcement learning (DRL) learning-based methods have been used in the architectural design and product management of smart agriculture applications. For example; Wang \cite{bu2019smart} aimed to design an edge-cloud computing-based smart agriculture system using DRL. This system consists of an agricultural data collection layer, edge computing layer, data transmission layer, and cloud computing layer. While data is collected from various sensors placed in the agricultural data collection layer, DRL is deployed in the cloud computing layer for instant smart decision-making. In another study, Alonso et al. \cite{alonso2020deep} proposed an architecture called Global Edge Computing Architecture (GECA) to realize Edge-IoT applications.  In this architecture, they developed SDN/NFV capabilities and a Double Deep-Q Learning model to control virtual data flows. Din et al. \cite{din2022deep} focused on the problem of monitoring crop health in a semi-structured farm. For this problem, a non-uniform area coverage algorithm based on the Double Deep Q-Network (DDQN) algorithm was developed.

On the other hand, studies on the explainability of smart agriculture practices have been quite limited. Sabrina et al. \cite{sabrina2022interpretable} designed an ML and fuzzy logic-based smart agriculture system aiming to increase crop production. The system is based on SVR, KNN and Naive Bayes models to monitor soil properties, and weather conditions, and detect possible abnormal conditions for the target crop. In the paper, an approach based on fuzzy rules is presented to ensure user trust and interoperability. In a similar study, Cartolano et al. \cite{cartolano2022explainable} focused on crop recommendation and effective feature extraction based on explainable ML methods in smart agriculture. They applied Extreme Gradient Boosting, Multi-layer Perceptron, and SVR models on the crop recommendation database. In these models, they obtained visualization with various graphics using SHAP and LIME, which are explainable artificial intelligence methods. 

Considering the existing gaps and shortcomings in the literature, in this work we present an explainable and interpretable decision support system based on IoT edge computing in the agricultural domain. The key contributions of this paper are summarized as follows.

\begin{itemize}
\item We propose a XAI based crop recommendation system called AgroXAI that provides a dynamic regional crop diversity.
\item We introduce a conceptual IoT architecture to operate the proposed system at the edge based on an edge computing approach. 
\item We also provide SHAP, LIME and ELI5 based post-hoc explainability of why the preferred crop is planted for an agricultural area.
\end{itemize}
This paper is structured as follows: Section \ref{Sec2_Agro} explains the proposed crop recommendation system model, used ML models and XAI methods. Section \ref{Sec3_ER} presents experimental result of ML models and discusses the XAI method results. Section \ref{Sec4_disc} provides a comprehensive discussion of security, privacy, ethical considerations, economic feasibility of the model and local preferences. Finally, Section \ref{Sec5_conc} concludes the paper.

\section{An Edge Computing-Based Explainable Crop Recommendation System (AgroXAI)}
In this section, we introduce the proposed system model for crop recommendation system and used ML and XAI methods.
\label{Sec2_Agro}

\begin{figure*}
    \centering
   \includegraphics[scale=.62]{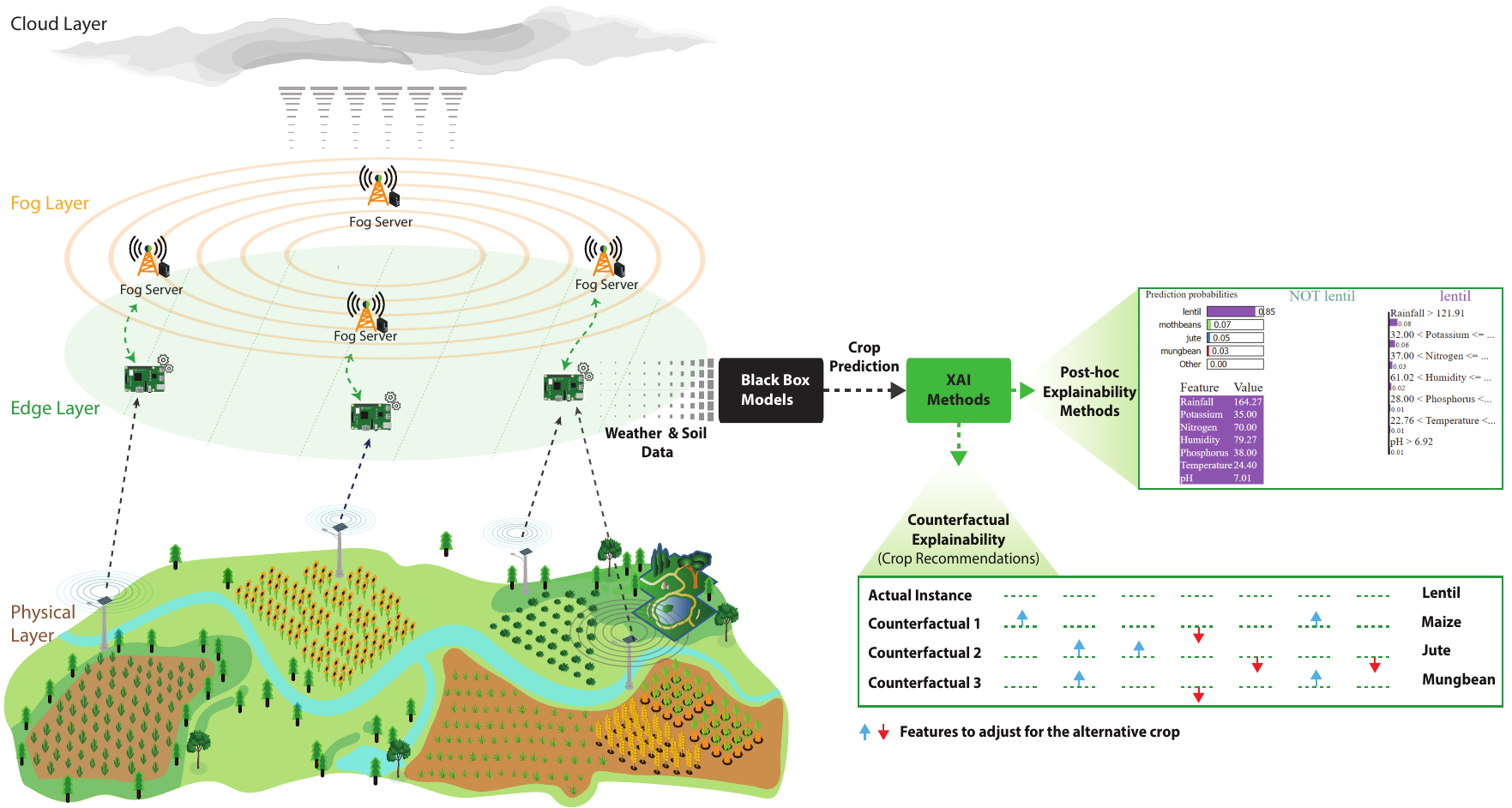}
    \caption{Proposed Edge Computing-Based Explainable Crop Recommendation System (AgroXAI)}
    \label{agroXAI system}
\end{figure*}

\subsection{System Model}
Next generation agriculture aims to make agricultural production more efficient, sustainable and predictable through technologies such as digitalization, automation and artificial intelligence. With this new vision of agriculture, data-driven decision-making processes will enable farmers to make more informed and region-specific crop choices. However, due to differences in climate, soil structure, water resources, temperature and humidity in different regions, there is a need for low-cost and effective decision support systems that can make localized decisions at the most extreme regional level. Therefore, in this study, we propose a data-driven and AI-supported crop recommendation system that will make regional crop selection more effective at the edge close to the end user. Specifically, the proposed system shown in Figure \ref{agroXAI system} is based on a conceptual IoT architecture consisting of physical layer, edge layer, fog layer and cloud layer. 
\begin{itemize}
    \item  Physical Layer: This layer includes sensors that measure the region's climate, soil structure, water resources, temperature and humidity, and actuators that provide conditions that can be changed in the region. In this way, it will be possible to periodically measure the current conditions for crop forecasting and change the environment when necessary.
    \item Edge Layer: At this layer, for each geographic region, there are end devices to analyze the locally collected data. these devices are capable of running classical ML and XAI methods (such as Raspberry Pi). Regionally, the edge devices are tasked with taking the data in their coverage area, predicting the appropriate crops for each region and providing an explanation of the prediction results. The ML models and XAI methods used in this study operate at this layer.
    \item Fog Layer: This layer includes hardware clouds that manage data traffic between the edge and the cloud layer and have the potential to provide network control.
    \item Cloud Layer: This layer includes resource-rich network devices that can perform computation and storage tasks that cannot be performed on edge systems in the proposed architecture. In particular, it collects all agricultural data in one center and provides batch analysis when needed.
\end{itemize}
In the physical layer of AgroXAI, farmers in each geographical region will be able to collect regional conditions at the time of planting crops with IoT sensors and see the most suitable crops to be planted according to ML and XAI results on edge devices. When needed, the obtained information can be stored on the fog server or cloud server and stored on central servers.

\begin{table*}
    \centering
    \caption{ML Model Parameters}
    \label{tab:mlParams}
    \resizebox{\textwidth}{!}{%
    \begin{tabular}{lll}
    \hline
       \textbf{Model}  & \textbf{Training Parameters} & \textbf{Best Parameters} \\
    \hline
       \textbf{KNN}  & \begin{tabular}[c]{@{}l@{}}\textit{n\_neighbours}: (10-50), \textit{metric}: (`euclidian', `cityblock')\end{tabular}   & \begin{tabular}[c]{@{}l@{}} \textit{n\_neighbours}: 11, \textit{metric}: `cityblock' \end{tabular} \\ \hline
        
       \textbf{RF}  & \begin{tabular}[c]{@{}l@{}}\textit{max\_depth}: (5-10), \textit{n\_estimators}: (10-150), \textit{criterion}: (`gini', `entropy') \end{tabular} & \begin{tabular}[c]{@{}l@{}l@{}} \textit{max\_depth}: 9,  \textit{n\_estimators}: 89, \textit{criterion}: `entropy' \end{tabular} \\ \hline
       \textbf{DT}  & \begin{tabular}[c]{@{}l@{}}\textit{max\_depth}: (10-150), \textit{criterion}: ('gini', `entropy'), \\ \textit{splitter}: (`best', `random'), \textit{min\_samples\_split}: [ 2, 4, 6, 8, 10, 12] \end{tabular}  &  \begin{tabular}[c]{@{}l@{}}\textit{max\_depth}: 131, \textit{criterion}: `gini', \\ \textit{splitter}: `best', \textit{min\_samples\_split}: 4 \end{tabular} \\ \hline
       \textbf{SVM}  & \begin{tabular}[c]{@{}l@{}} \textit{kernel}: `rbf', 
       \quad \textit{C}: [0.001, 1], \textit{gamma}: [0.01, 0.1] \\ \textit{kernel}: `linear', 
       \quad \textit{C}: [0.001,0.01, 0.1] \end{tabular} & \begin{tabular}[c]{@{}l@{}} \textit{kernel}: `linear' 
       \quad \textit{C}: 0.01 \end{tabular}\\ \hline
       \textbf{LGBM}  & \begin{tabular}[c]{@{}l@{}}\textit{num\_leaves}: (5-20),  \textit{learning\_rate}: [0.1, 0.01, 0.001],  \\ \textit{n\_estimators}: (10-50) \end{tabular} & \begin{tabular}[c]{@{}l@{}} \textit{num\_leaves}: 5,  \textit{learning\_rate}: 0.1, \textit{n\_estimators}: 43 \end{tabular} \\ \hline
       \textbf{MLP}  & \begin{tabular}[c]{@{}l@{}}\textit{activation}: (`tanh', `relu'),  \textit{learning\_rate}: (`constant', `adaptive'), \\\textit{solver}: (`sgd', `adam'), 
       \textit{alpha}: [0.0001, 0.001,0.1,0.2,0.3,0.4,0.5], \\
       \textit{hidden\_layer\_sizes}: [(10,10),(10,20), (10,30), \\
       \quad (10,40),(10,30,10),(10, 30, 50, 25)] \end{tabular} &  \begin{tabular}[c]{@{}l@{}}\textit{activation}: `relu',   \textit{learning\_rate}: `constant',\\ \textit{solver}: `adam', \textit{alpha}: 0.5, \\
       \textit{hidden\_layer\_sizes}: (10, 30, 50, 25) \end{tabular} \\
    \hline
    \end{tabular} }
\end{table*}

\subsection{ML Models}
In this paper, we used K-Nearest Neighbors (KNN), Random Forest (RF), Decision Tree (DT), Support Vector Machine (SVM), LightGBM (LGBM) and Multilayer Perceptron (MLP) models to classify the crop. Table \ref{tab:mlParams} shows the parameters used in the training of these models.

\subsubsection{KNN}
KNN, which is based on the distance between data points, is used to solve classification and regression problems. It uses the nearest "k" pieces of data to decide which class the data belongs to. The success of KNN is determined by the number of neighbors (k) and the distance calculation algorithm used \cite{han}.
\subsubsection{RF}
RF is an ensemble approach. A forest is made up of many independent decision tree classifiers. Changing the attribute choices at random instead of the training examples is the main concept. Every tree in the forest generates a result by doing its own assessment. Each tree votes, and the value with the highest score is chosen \cite{han}.
\subsubsection{DT}
DT is a model that predicts by learning simple rules extracted from data features, represented by a tree representation. Instances are ordered from root to leaf node. Each node in the tree indicates a feature of the instance to be tested. Based on the tested features, the path from the root to a leaf node leads to the class label. The overfitting problem in DT can be encountered. Pruning can be applied to solve this problem \cite{michael2011} .
\subsubsection{SVM}
SVM is a method for classifying both linear and non-linear data. In this method, data are classified into predefined classes with the help of a hyperplane. In determining the hyperplane, margins and support vectors, which are data points close to the hyperplane, are used\cite{han}. 
\subsubsection{LGBM}
LGBM is a type of histogram-based boosting method. It is capable of rapidly processing big datasets \cite{ke2017lightgbm}.  It uses the leaf-wise strategy, one of the methods used in learning decision trees. This strategy allows an unbalanced tree to grow and aims to split the leaf with the most loss. In this way, the loss is minimized \cite{shi2007best}.
\subsubsection{MLP}
MLP is a simple feedforward neural network and basically consists of three layers. The first layer is the input layer, which accepts incoming signals and then passes them on to neurons in the hidden layer. Calculations are performed in the hidden layer. There may be one hidden layer, or there could be multiple hidden layers. The last layer, the output layer, accepts the output signals from the hidden layer and creates the output pattern of the network \cite{michael2011}.

\subsection{Explainable AI Methods}
In the classification process, not only is it important for models to achieve high accuracy, but the interpretability of the models' decisions is also of critical importance. Providing interpretability in these developed systems emerges as a factor that enhances trust in the model decisions \cite{gurbuz2023explainable}. Therefore, in this study, we employ IML methods such as ELI5, LIME, SHAP, and Counterfactual.

\subsubsection{Explain Like I'm 5 (ELI5)}
 ELI5 is a Python package used for the explainability of black box models. It provides both global and local explainability. ELI5 uses tree models for calculating feature weights. The contribution of the feature to the decision is based on how much the score has changed from parent to child at each node of the tree.  \cite{bhattacharya2022}

\subsubsection{SHapley Additive exPlanations (SHAP)}
SHAP is an explainability method based on game theory. In this method, a value called Shapley value is calculated for each feature, which expresses the contribution of the feature to the outcome. Shap provides both local and global explanation \cite{molnar2022}.

\subsubsection{Local Interpretable Model-agnostic Explanations (LIME)}
 LIME method examines how the model works by changing the inputs and observing how the predictions vary. LIME is model-agnostic and provides local explanations \cite{ribeiro2016should}.

\subsubsection{Counterfactual}

Counterfactual is a human-friendly explainability method that explains the smallest change in feature values and transforms the prediction into a predefined output. Although a counterfactual tries to generate feature values as close as possible to the corresponding instance, it is not always possible to find a counterfactual for every predefined prediction \cite{molnar2022}. 

\section{Experimental Results and Evaluation}
\label{Sec3_ER}
\subsection{Dataset Information}
In this paper, we use the agriculture dataset available in Kaggle \cite{Chitrakumari_2022_kaggle}. The dataset contains 2200 rows of data and seven features. The features are presented in Table \ref{tab:featsAndDesc}. In addition, the dataset contains 22 different crop types as target labels. 
\begin{table}[!htb]
    \tiny
    \centering
    \caption{Dataset Features and Descriptions}
    \label{tab:featsAndDesc}
    \resizebox{0.8\columnwidth}{!}{%
    \begin{tabular}{ll}
    \hline
       \textbf{Features}  & \textbf{Descriptions} \\
    \hline
       Nitrogen  &  Amount of Nitrogen in soil \\ 
       Phosphorus  & Amount of Phosphorus in soil \\ 
       Potassium  & Amount of Potassium in soil \\ 
       Temperature  & The average soil temperatures \\
       Humidity  & Amount of humidity \\ 
       ph  & pH level of the soil \\ 
       Rainfall  & Amount of rainfall \\ 
    \hline
       Target  & Types of crop \\ 
    \hline
    \end{tabular}}
\end{table}
These crops are \textit{apple, banana, blackgram, chickpea, coconut, coffee, cotton, grapes, jute, kidney beans, lentil, maize, mango, moth beans, mungbean, muskmelon, orange, papaya, pigeon peas, pomegranate, rice} and  \textit{watermelon.}

\subsection{ML Results and Evaluation} 
We tested all used models with the best parameters in Table \ref{tab:mlParams} and obtained the results in Table \ref{tab:mlClassResults}. 
When comparing classification models based on precision, recall, F1 score, and accuracy, all models show strong performance with an overall value above 95\%. In particular, RF outperforms the others with outstanding precision, recall, F1 score, and accuracy reaching 99.24\%. The DT follows closely behind with a balanced performance with an accuracy of 98.48\%. KNN, SVM, LGBM, and MLP also show strong performances, reaching over 95\% accuracy. The success of RF here may be due to combining multiple decision trees and mitigating overfitting through random sampling, which tends to provide robust generalization.

\begin{table}[!htb]
    \caption{Classification Results of ML Models}
    \label{tab:mlClassResults}
    \resizebox{\columnwidth}{!}{%
    \begin{tabular}{lllll}
        \hline
        \textbf{} & \textbf{Precision} & \textbf{Recall} & \textbf{F1-Score} & \textbf{Accuracy} \\ \hline
        \textbf{KNN} & 97.1071 & 96.6667 & 96.6198 & 96.6667 \\ \hline
        \textbf{RF} & 99.3395 & 99.2424 & {\textbf{99.2312}} & {\textbf{99.2424}} \\ \hline
        \textbf{DT} & 98.5620 & 98.4848 & {\textbf{98.4742}} & {\textbf{98.4848}} \\ \hline
        \textbf{SVM} & 97.7694 & 97.4242 & 97.4163 & 97.4242 \\ \hline
        \textbf{LGBM} & 97.7930 & 97.5758 &\textbf{97.5527} & \textbf{97.5758} \\ \hline
        \textbf{MLP} & 95.7698 & 95.6061 & 95.5945 & 95.6061 \\ \hline
    \end{tabular}}
\end{table}

\begin{figure}[!b]
    \centering
    \includegraphics[width=3.4in]{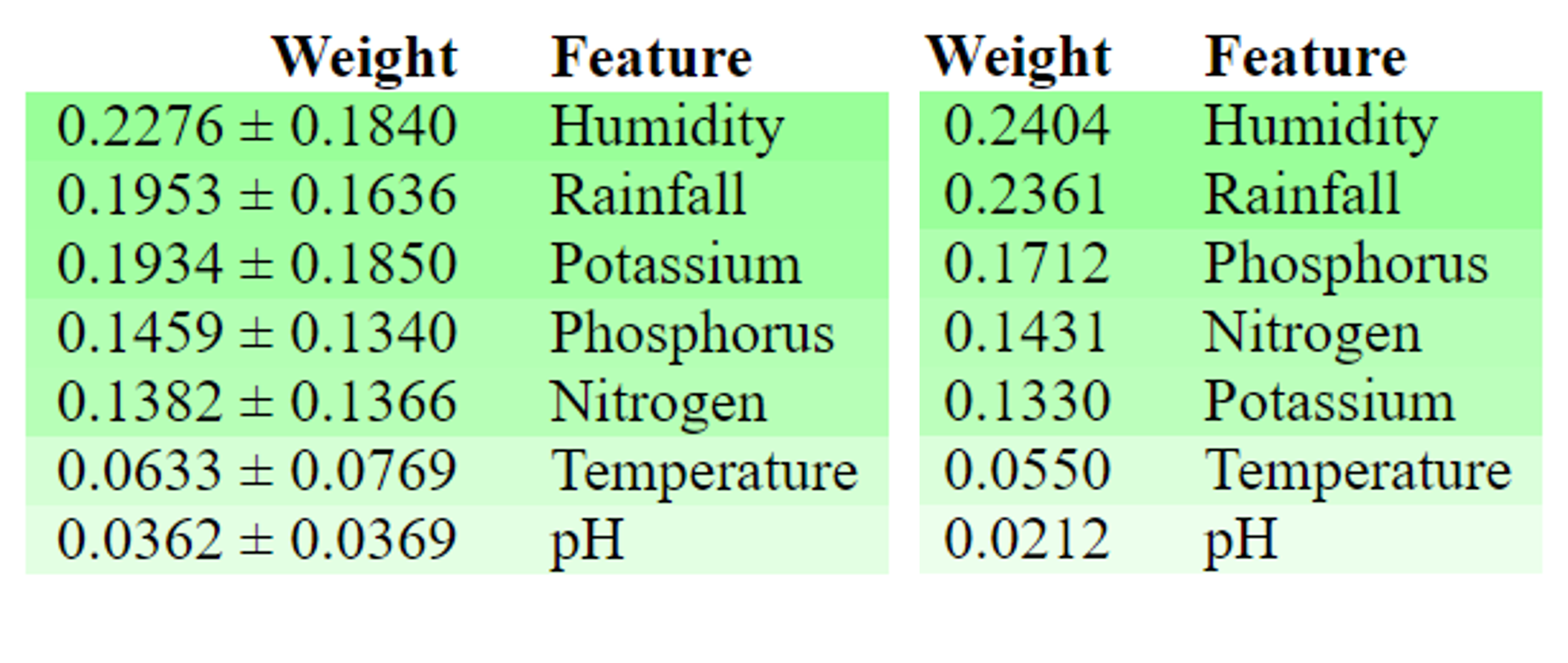}
    \caption{ELI5 global explanations for RF (Left) and LGBM (Right)}
    \label{fig:Eli5Global}
\end{figure}
\begin{figure*}
   \centering
   \includegraphics[width=5.6in]{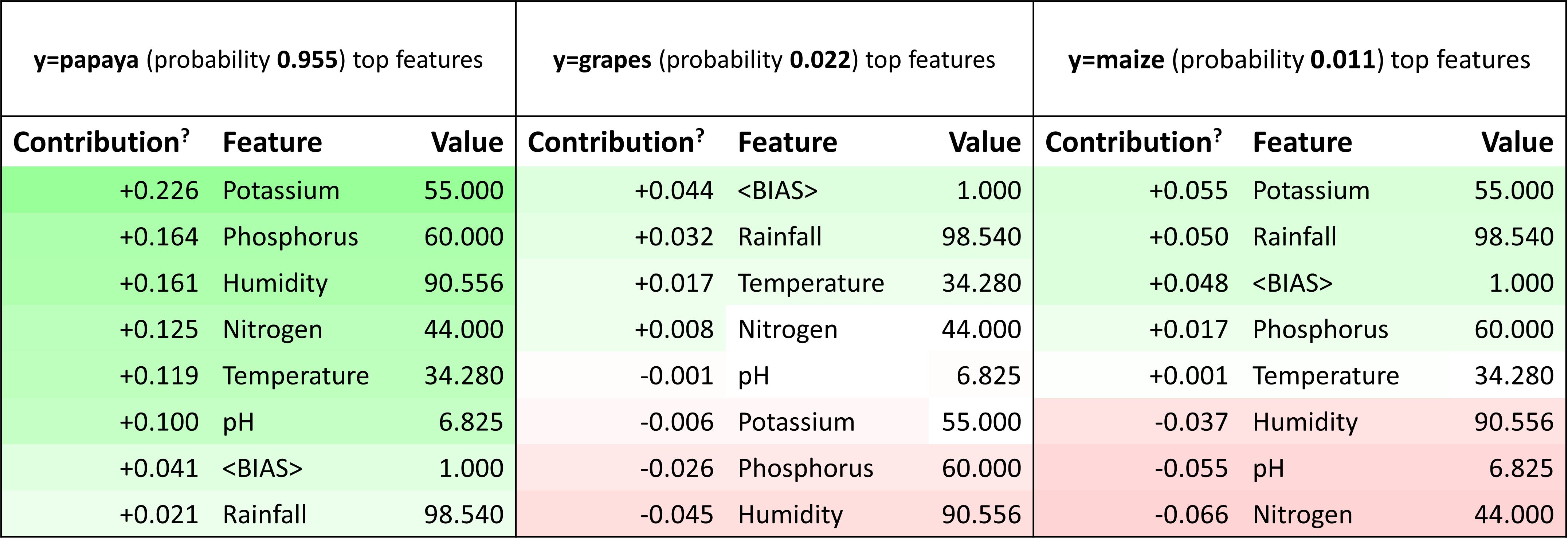}
   \caption{ELI5 local explanations for RF (Randomly selected sample test data: {Nitrogen = 44, Phosphorus = 60, Potassium = 55, Temperature = 34.28046,
Humidity = 90.555618, pH = 6.825371,
Rainfall = 98.540474})}
   \label{fig:rfEliLocal3}
   \end{figure*}
\begin{figure*} 
    \centering
    \includegraphics[width=5.6in]{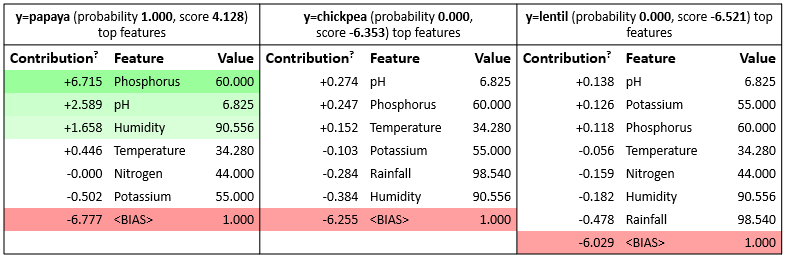}
    \caption{ELI5 local explanations for LGBM (Randomly selected sample test data: {Nitrogen = 44, Phosphorus = 60, Potassium = 55, Temperature = 34.28046,
Humidity = 90.555618, pH = 6.825371,
Rainfall = 98.540474})}
    \label{fig:lgbmEliLocal3}
\end{figure*}

\subsection{XAI Results and Evaluation}
We used ELI5, SHAP, and Counterfactual explainability methods to explain ML models developed for crop diversity. In this context, we first used ELI5 and SHAP methods for local and global explanation of the decisions of the most successful ML models RF and LGBM models. Then, we determined the crop varieties to be grown in the relevant region using the counterfactual method.

The global explainability of the ELI5 method with respect to RF and LGBM models is shown in Figure \ref{fig:Eli5Global}. These graphs show the most effective attributes for classifying the crop variety according to the weather and soil conditions in the region. According to the information in the figure, it is seen that the first two features that contribute the most to the decision in crop classification for RF and LGBM models are \textit{``Humidity''} and \textit{``Rainfall''}. The third attribute is \textit{``Potassium''} for RF and \textit{``Phosphorus''} for LGBM. This information shows that moisture and rainfall features are critical in determining crop diversity.

For local interpretability, a single data instance is selected, and the model's decision output for this data is analyzed. In this context, the local interpretability of the most successful models, RF and LGBM, has been extracted for randomly selected data instances. The local interpretability of RF and LGBM models for randomly selected data are given in Figure \ref{fig:rfEliLocal3} and Figure \ref{fig:lgbmEliLocal3}, respectively. In Figure 2, the RF model predicted the selected single data instance as papaya with a probability of 95.5\%, grapes with 2.2\%, and maize with 1.1\%. The interpretability of this result was derived using the ELI5 method. According to this, “Potassium”, "Phosphorus" and "Humidity" were the most effective features for the papaya decision. On the other hand, "Rainfall", "Potassium" and "Temperature" were more effective for maize and grape predictions. For the same test data, the LGBM model predicted papaya with 100\% accuracy. The interpretability of this result was obtained using the ELI5 method and is shown in Figure 3. According to this, it was determined that the most effective features in the decision of the LGBM model were Phosphorus, pH and Humidity", respectively. When both global and local explanation results of the ELI5 method are evaluated together, it is observed that the values of humidity, rainfall, phosphorus, and potassium are decisive features in regional crop production.

\begin{figure*}
    \centering
\includegraphics[width=7.3in]{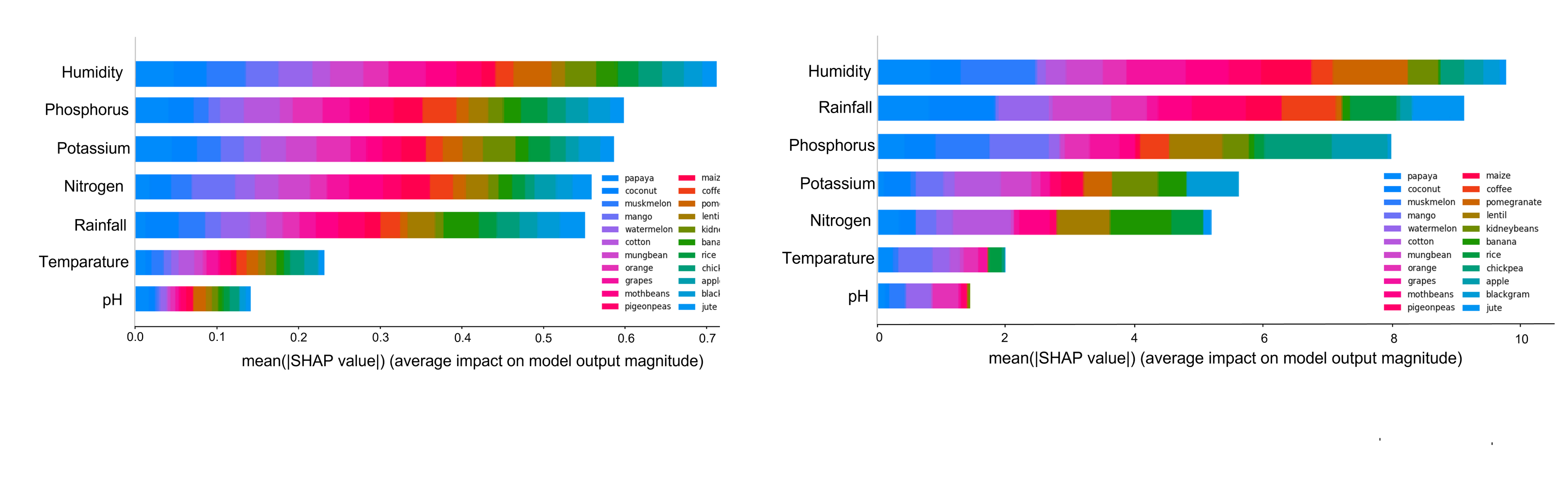}
    \caption{SHAP global explanations for RF (Left) and LGBM (Right)}
    \label{fig:ShapGlobal}
\end{figure*}

In this study, more than one method is used for better interpretability of model decisions. In this context, RF and LGBM model decisions are also explained both locally and globally with the SHAP method. First, the global explainability of the SHAP method for the RF and LGBM models is shown using the summary plot in Figure \ref{fig:ShapGlobal}. The features are ranked according to their importance. Here, each color represents the contribution to different crop types. For both the RF and LGBM models, "Humidity" is identified as the most important feature for crop classification. While the other decisive features for the RF model are phosphorus, potassium, and nitrogen, they are ranked as rainfall, phosphorus, and potassium for the LGBM model. 

\begin{figure*}[ht]
\centerline{\includegraphics[width=6.6in]{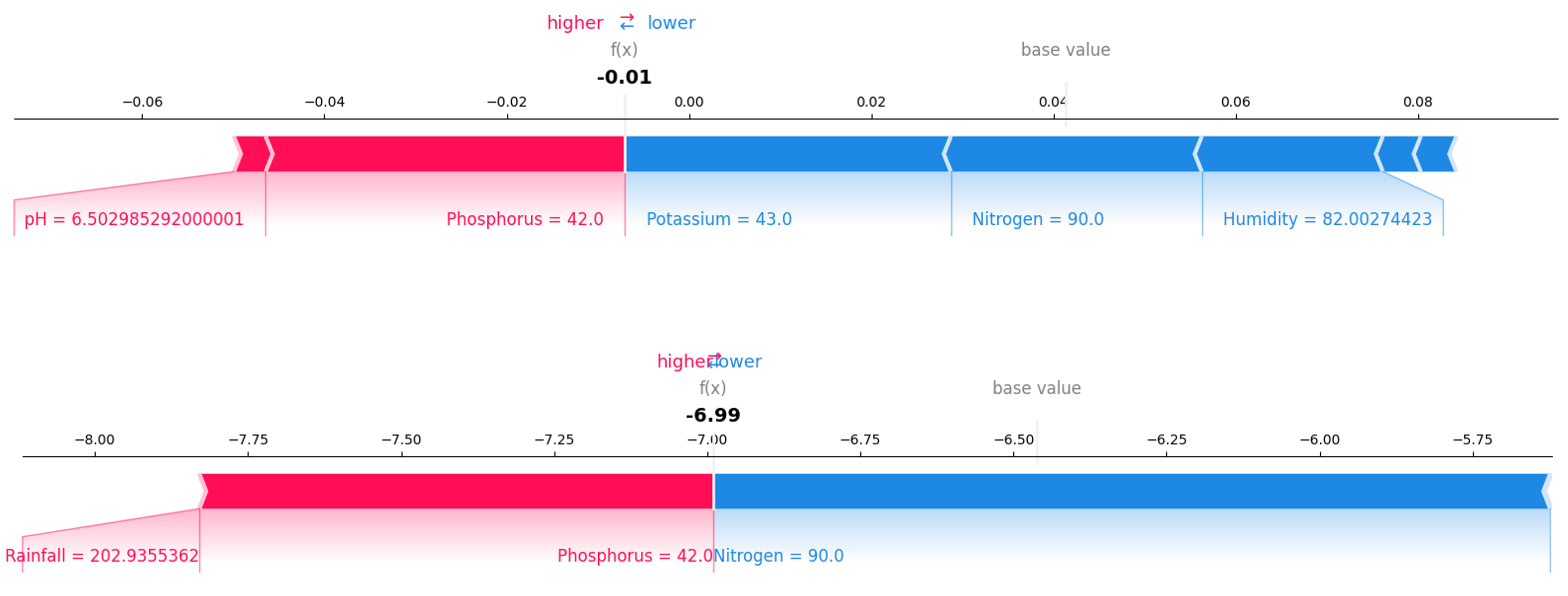}}
\caption{SHAP local explanations for RF (Top) and LGBM (Bottom)- Randomly selected sample data: Nitrogen = 39, Phosphorus = 77, Potassium = 21, Temperature = 22.997744, Humidity = 60.242188, pH = 4.603563, Rainfall = 159.689346}
\label{fig:rfLgbmShapLocal28}
\end{figure*}

Then, SHAP local explainability results were obtained for the sample test data provided in the description of Figure \ref{fig:rfLgbmShapLocal28} and illustrated using a force plot. In the force plot, the feature occupying the largest area is the one contributing the most to the decision. Accordingly, "Phosphorus" was the most contributing feature for the RF model, while "Nitrogen" played the same role for the LGBM model. Features shown in red have a positive contribution to the decision, whereas those in blue have a negative contribution. For the RF model, "Phosphorus" and "pH" contributed positively, while "Potassium," "Nitrogen," and "Humidity" contributed negatively. In the LGBM model, "Phosphorus" and "Rainfall" contributed positively, whereas the most influential feature, "nitrogen," had a negative contribution.

\begin{figure*}
    \centering
    \includegraphics[width=7in]{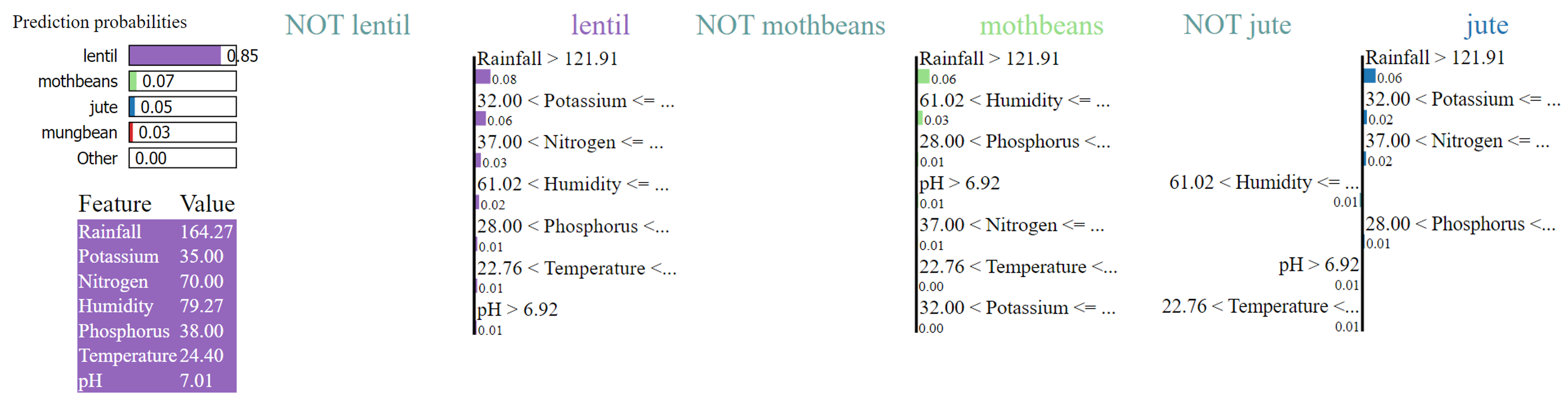}
    \caption{LIME local explanations for RF (Used sample data: Nitrogen = 70, Phosphorus = 38, Potassium = 35, Temperature = 24.397362, Humidity = 79.268616, pH = 7.014064, Rainfall = 164.269699 )}
    \label{fig:rfLime13}
\end{figure*}

We also examined the local explanations of the RF and LGBM model according to LIME, another local explainability method. As shown in Figure \ref{fig:rfLime13}, for the randomly selected data, the RF model predicted lentil with 85\% probability, mothbean with 7\% probability, jute with 5\% probability and mungbean with 3\% probability. The contribution of the features for this prediction is shown on the right side of the figure from maximum to minimum. Rainfall greater than 121.91, Potassium greater than 32 and Nitrogen greater than 37 led to the conclusion of lentils. Rainfall greater than 121.91 and Humidity content greater than 61.02 contributed to the decision for mothbean. For jute, Rainfall greater than 121.91, Potassium greater than 32, Nitrogen greater than 37 and Phosphorus greater than 28 contributed positively, while Humidity greater than 61.02, pH greater than 6.92 and Temperature greater than 22.76 negatively influenced the decision. 

\begin{figure*}
    \centering
    \includegraphics[width=7in]{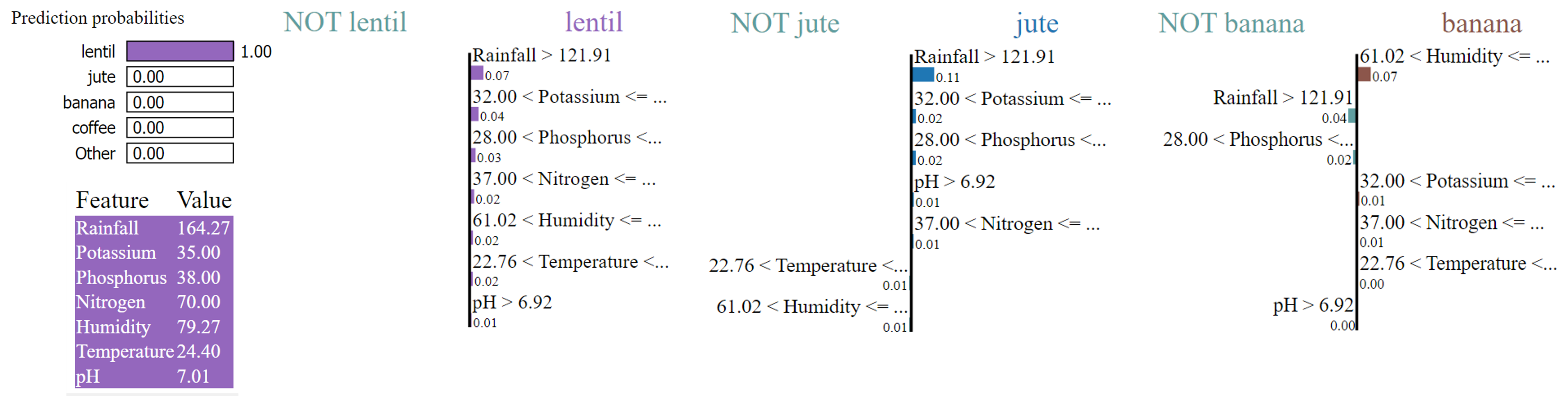}
    \caption{LIME local explanations for LGBM (Used sample data: Nitrogen = 70, Phosphorus = 38, Potassium = 35, Temperature = 24.397362, Humidity = 79.268616, pH = 7.014064, Rainfall = 164.269699 )}
    \label{fig:lgbmLIME}
\end{figure*}

On the other hand, the LGBM model classified the same sample data as lentil with 100\% probability as shown in Figure \ref{fig:lgbmLIME}. This decision is largely influenced by rainfall and potassium features. 

\begin{table*}
    \centering
    \caption{Counterfactuals for RF}
    \label{tab:counterfactualsRF}
    \begin{tabular}{lllllllll}
        \hline
        \textbf{Type of Instance} & \textbf{Nitrogen} & \textbf{Phosphorus} & \textbf{Potassium} & \textbf{Temperature} & \textbf{Humidity} & \textbf{pH} & \textbf{Rainfall} & \textbf{Label} \\ \hline
        Actual Instance & 44 & 60 & 55 & 34.28046 & 90.555618 & 6.825371 & 98.540474 & Papaya \\
        Counterfactual-1 & 117 & 60 & 55 & 34.281461 & 45.50041 & 6.825371 & 98.550477 & Banana \\
        Counterfactual-2 & 44 & 60 & 38 & 34.281461 & 60.18227 & 6.825371 & 98.550477 & Mango \\
        Counterfactual-3 & \textbf{85} & 60 & 55 & 34.281461 & \textbf{85.29596} & 6.825371 & \textbf{295.154486} & Rice \\ \hline
    \end{tabular}
\end{table*}

\begin{table*}
    \centering
    \caption{Counterfactual Explainability for LGBM}
    \label{tab:counterfactualsLGBM}
    \begin{tabular}{lllllllll}
        \hline
        \textbf{Type of Instance} & \textbf{Nitrogen} & \textbf{Phosphorus} & \textbf{Potassium} & \textbf{Temperature} & \textbf{Humidity} & \textbf{pH} & \textbf{Rainfall} & \textbf{Label} \\ \hline
        Actual Instance & 44 & 60 & 55 & 34.28046 & 90.555618 & 6.825371 & 98.540474 & Papaya \\
        Counterfactual-1 & 93 & 86 & 55 & 34.281461 & 90.655616 & 5.916632 & 98.550477 & Banana \\
        Counterfactual-2 & 137 & 60 & 55 & 34.281461 & 45.35615 & 6.825371 & 98.550477 & Mango \\
        Counterfactual-3 & 44 & 60 & 55 & 34.281461 & 29.08982 & 6.825371 & 259.863518 & Rice \\ \hline
    \end{tabular}
\end{table*}

Up to this point, we have evaluated the post-hoc explainability of the model results by using ELI5, SHAP and LIME. In this study, we also aimed to provide alternative crop recommendation based on edge computation regionally in addition to the interpretability of the current decisions. In this context, we used the counterfactual explainability method to identify the list of other crops that can be planted for each predicted crop. In counterfactual explainability, each selected data sample is referred to as the “actual instance,” and the class for selected data is predicted. In addition to this result, alternative counterfactual suggestions are provided for the output. In this way, regionally appropriate products and alternative suggestions are obtained. Based on this information, counterfactual explainability results for randomly selected data samples are presented in Table IV and Table V.

\begin{figure*}[!t]
    \centering
    \begin{subfigure}{0.49\textwidth}
        \centering
        \includegraphics[width=\linewidth]{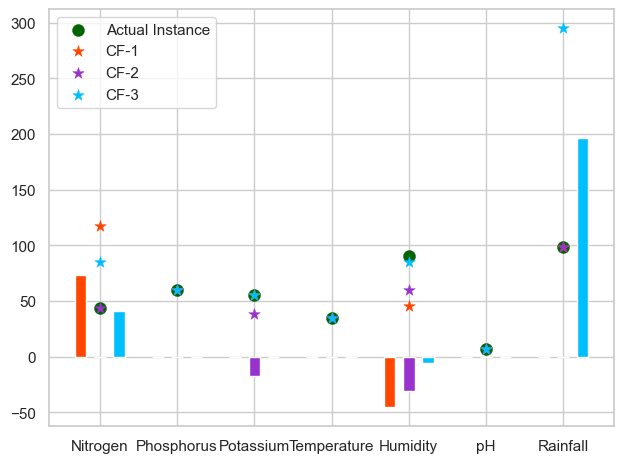}
        \caption{Counterfactuals for RF}
        \label{fig:fig1}
    \end{subfigure}
    \hfill
    \begin{subfigure}{0.49\textwidth}
        \centering
        \includegraphics[width=\linewidth]{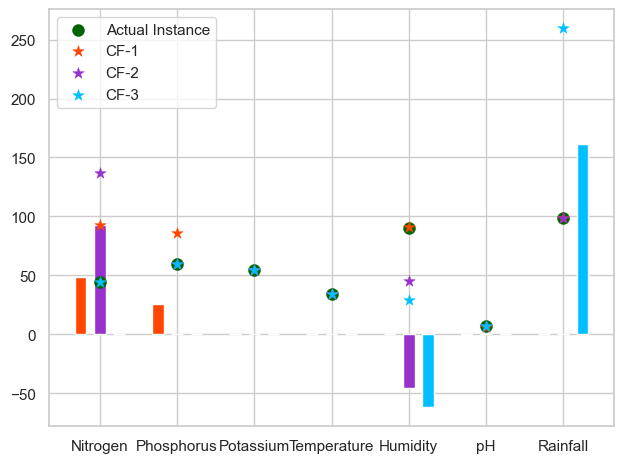}
        \caption{Counterfactuals for LGBM}
        \label{fig:fig2}
    \end{subfigure}
    \caption{Counterfactual Explainability of RF and LGBM models}
    \label{fig:figures}
\end{figure*}

As seen in Table 4, the RF model has predicted the selected actual instance data as Papaya. However, with some changes in the input data features, it is observed that counterfactually, the instance could also be classified as Banana, Mango, or Rice. However, it is noted that for these alternative crops, changes in values such as rainfall, nitrogen, and humidity are required. These value changes and their directions are illustrated in Figure \ref{fig:fig1}. In Figure \ref{fig:figures}, the bar graphs up and down from 0 show the changes in the feature values for the counterfactual classes. For instance, a farmer who wants to grow Rice in a region instead of Papaya as given in counterfactual-3 would need to increase the Nitrogen and Rainfall values and decrease the Humidity values in that region. However, it may not be inherently possible to change some characteristics. For this reason, it would be appropriate to turn to crops with features that can be changed.

In Table \ref{tab:counterfactualsLGBM}, the actual instance and counterfactual outputs of the LGBM model for the same test data are provided. Figure \ref{fig:fig2} illustrates the changes for each counterfactual output compared to the actual instance in the LGBM model. According to this result, a farmer who wants to grow Rice given in counterfactual-3 instead of Papaya should expect an increase in Rainfall and a decrease in Humidity in the region.

When all the explainability method results are evaluated, post-hoc explainability of the ML models has been achieved through methods such as ELI5, SHAP, and LIME. Accordingly, although each algorithm highlights different features in different orders globally and locally, similar features were emphasized in the decision-making process. This phenomenon corresponds to the disagreement problem in the field of XAI. While the ML models produce the same predictions, they may provide different explanations regarding the impact of features on the outcome. The results obtained in this study confirm this observation.  On the other hand, we used counterfactual explainability to provide product suggestions for growing alternative crops and diversification. The results of this methodology also yielded valuable results in terms of providing alternative products regionally. 

\section{Discussion} \label{Sec4_disc}
The AgroXAI system ensures model explainability while fostering crop diversity by offering locally relevant alternative crop recommendations. These features allow for the transparent disclosure of the rationale behind agricultural system decisions, thereby enhancing the decision-making process of farmers. While the AgroXAI system offers numerous advantages, it is essential to also address critical aspects such as security, privacy, ethical considerations, economic feasibility, and local preferences. Therefore, this section will explore these key issues in detail.

In explainable systems, there is generally a trade-off between accuracy and interpretability. Accordingly, models with high interpretability tend to have lower accuracy, while models with higher accuracy may exhibit reduced interpretability. In the context of AgroXAI, this highlights the need to strike a balance between optimizing the accuracy and ensuring transparency for users. Therefore, the selection of XAI methods in systems like AgroXAI may depend on a choice between maximizing model accuracy and presenting the system's decisions in an understandable and user-friendly manner. 

While the AgroXAI system facilitates transparency and informed decision-making, its reliance on user data and sensor-derived information introduces potential concerns. If sensor security is not adequately ensured, malicious actors could create privacy and security risks. As such, addressing security and privacy is paramount in the design and implementation of the system to ensure its trustworthiness.

In another dimension, AgroXAI aims to minimize environmental impacts by helping farmers optimize resource usage. By employing the counterfactual method, it provides valuable information on how farmers should adjust water and fertilizer levels for different crops, thus enhancing productivity. For instance, this feature supports sustainable agriculture by promoting the efficient use of water resources and preserving soil health through optimized fertilizer use. However, the potential impacts of crop recommendations on livelihoods and food security must also be carefully considered. For instance, if the recommended crops are not suitable for local climate conditions or soil characteristics, crop failure may occur, leading to a significant decline in farmers' income. Such losses present considerable risks, particularly in regions where agriculture and food security are critical. In light of these challenges, the ability of developed systems to adapt to changing environmental conditions and provide appropriate recommendations becomes increasingly important, especially in a context where climate variability and soil degradation are escalating.

\section{Conclusion}
\label{Sec5_conc}
In this study, we propose a digitalized, remotely manageable, data-driven, and AI-supported system aimed at achieving the sustainability, efficiency, capacity enhancement, and smart agricultural production goals targeted by Agriculture 4.0. We present an explainable AI-powered smart crop selection and recommendation system to meet these needs. The proposed system is structured to identify crops suitable for specific geographical conditions and to provide a list of alternative crops. By focusing on the promising area of XAI within the field of AI, AgroXAI has the potential to build trust on the user side. The obtained results have increased the interpretability of the model decisions of the developed system through different XAI methods and established confidence among users. Additionally, the results provide insights into the environmental factors that need to change for regional crop alternatives. In this respect, AgroXAI is expected to contribute to customized production processes by serving as both an intelligent and explainable decision support system in agricultural production. In the coming years, with the widespread adoption of such systems, farmers will be able to better understand region-specific conditions, increase efficiency, minimize environmental impacts, and optimize product diversity.

\bibliographystyle{IEEEtran}
\bibliography{refs}

\end{document}